\documentclass[akbc,twoside,11pt,lettersize]{article}
\usepackage{akbc}
\usepackage[table,xcdraw]{xcolor}
\usepackage{booktabs}
\usepackage{multirow}
\usepackage{tabularx}
\usepackage{amsmath}
\usepackage{amssymb}
\usepackage{booktabs}
\usepackage{graphicx}
\usepackage{subcaption}
\usepackage[table,xcdraw]{xcolor}
\usepackage{float}
\restylefloat{table}

\usepackage{times}
\usepackage{latexsym}

\usepackage[T1]{fontenc}

\usepackage[utf8]{inputenc}

\usepackage{microtype}

\usepackage{amsmath}

\DeclareMathOperator*{\argmax}{arg\,max}

\akbcheading
\ShortHeadings{How Optimal is Greedy Decoding for Extractive Question Answering?}{Or Castel, Ori Ram, Avia Efrat \& Omer Levy}
\finalcopy 
\begin{document}

\title{How Optimal is Greedy Decoding for \\ Extractive Question Answering?}

\author{\name Or Castel$^1$ \email or.castel@cs.tau.ac.il \\ \name Ori Ram$^1$ \email ori.ram@cs.tau.ac.il \\ \name Avia Efrat$^1$ \email avia.efrat@cs.tau.ac.il \\ \name Omer Levy$^{1,2}$ \email levyomer@cs.tau.ac.il \\
       \addr $^1$Blavatnik School of Computer Science, Tel Aviv University\\
       $^2$Meta AI Research
       }


\maketitle

\begin{abstract}
Fine-tuned language models use greedy decoding to answer reading comprehension questions with relative success.
However, this approach does not ensure that the answer is a span in the given passage, nor does it guarantee that it is the most probable one. 
Does greedy decoding actually perform worse than an algorithm that \textit{does} adhere to these properties?
To study the performance and optimality of greedy decoding, we present \textit{exact-extract}, a decoding algorithm that efficiently finds the most probable answer span in the passage.
We compare the performance of T5 with both decoding algorithms on zero-shot and few-shot extractive question answering.
When no training examples are available, exact-extract significantly outperforms greedy decoding.
However, greedy decoding quickly converges towards the performance of exact-extract with the introduction of a few training examples, becoming more extractive and increasingly likelier to generate the most probable span as the training set grows.
We also show that self-supervised training can bias the model towards extractive behavior, increasing performance in the zero-shot setting without resorting to annotated examples.
Overall, our results suggest that pretrained language models are so good at adapting to extractive question answering, that it is often enough to fine-tune on a small training set for the greedy algorithm to emulate the optimal decoding strategy.\footnote{Our code and models are publicly available at: \url{https://github.com/ocastel/exact-extract}}
\end{abstract}

\section{Introduction}

Extractive question answering is the task of answering a question given a passage, assuming the answer appears as a span in the passage. 
It is a main component in state-of-the-art methods for open-domain question answering \cite{karpukhin-etal-2020-dense, izacard-grave-2021-leveraging}, and can facilitate numerous other NLP tasks, such as relation extraction \cite{levy-etal-2017-zero}, coreference resolution \cite{wu-etal-2020-corefqa}, named entity recognition \cite{li-etal-2020-unified}, and more.
Generative language models usually address this task via greedy decoding algorithms, which do not guarantee two key properties:
(1) they are not \textit{extractive}, i.e. they can produce texts that are not spans in the passage,
(2) they are not \textit{exact}, i.e. they do not necessarily generate the most probable output according to the model.
In this work, we show that despite lacking any formal guarantees, greedy decoding can approach the performance of a theoretically optimal decoding algorithm across a variety of extractive question answering benchmarks, even when only a few training examples are available.

\begin{figure}[t]
 \centering
 \includegraphics[width=0.5\columnwidth]{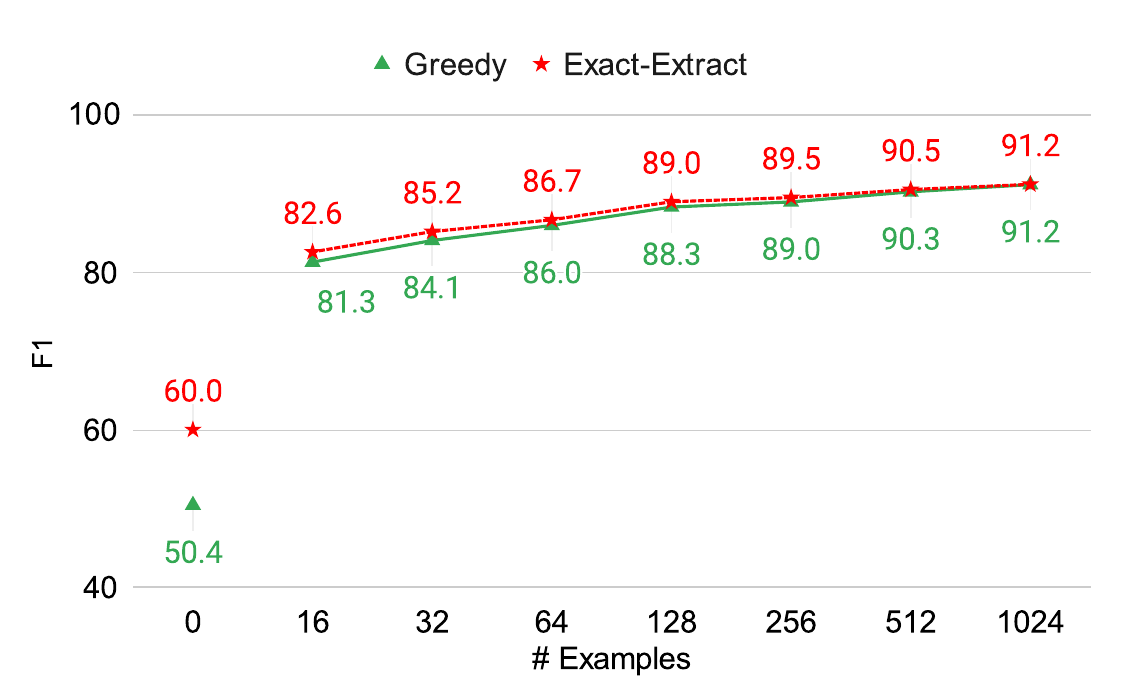}
 \caption{Performance of T5-large on SQuAD when using greedy (green) and optimal (red) decoding, given different amounts of training examples.
 As the amount of examples increases, the performance gap between the decoding algorithms diminishes.}
\label{fig:intro_fig}
\end{figure}

To that end, we introduce \textit{exact-extract}, a decoding algorithm that efficiently calculates the model-assigned probabilities of all spans in the passage, allowing us to (provably) select the most probable span.
We compare greedy decoding with exact-extract on the recently-proposed few-shot question answering benchmark \cite{ram-etal-2021-shot}, which contains subsampled training sets of 16 to 1024 examples from 8 different datasets.
Specifically, we fine-tune a pretrained language model, T5-large \cite{2020t5}, and measure the performance of both decoding algorithms. 

In the zero-shot setting, where no annotated examples  are available, there is a significant performance margin (11.3 F1 points on average across datasets) between greedy decoding and exact-extract.
This gap quickly shrinks as more annotated examples are introduced; even 16 training examples are enough to narrow the average performance margin to 2.8 points, with 1024 examples diminishing it to 0.3.
Figure~\ref{fig:intro_fig} shows this trend on the SQuAD dataset \cite{rajpurkar-etal-2016-squad}.

We further measure how often greedy decoding generates spans from the given passage (i.e. the algorithm's \textit{extractiveness}), and observe a strong correlation between extractiveness and performance.
In particular, we notice that in the zero-shot setting, where exact-extract strongly outperforms greedy decoding, the greedy algorithm is substantially less extractive.
To increase extractiveness, we propose an additional self-supervised pretraining phase inspired by recurring span selection \cite{ram-etal-2021-shot}.
Training with this objective enhances the model's tendency to generate answers from the context, and consequentially improves the performance of greedy decoding in this challenging setting.

Overall, our results demonstrate that although greedy decoding does not explicitly guarantee either extractiveness or exactness,
the underlying model (T5) adapts so well to the task of extractive question answering, that even a few examples are enough to allow the naive greedy decoding algorithm to generate answers that rival those of an optimal decoding algorithm.

\section{Problem Setting}

The task of \textit{extractive question answering} (extractive QA) \cite{rajpurkar-etal-2016-squad} is to select a span $a$ from a given passage $T$ that answers a question $Q$.
In this work, we focus on few-shot and zero-shot extractive QA \cite{ram-etal-2021-shot}, where the learner is given a small set of training and development examples (from 16 to 1024 examples), or none at all.
These settings resemble real-world use-cases, where an abundance of data is not necessarily available, as annotating large datasets is expensive and may even require domain and language experts. 

Extractive QA is typically modeled via span selection models that point to the start and end of the answer span \cite{seo2018bidirectional,devlin-etal-2019-bert}.
This approach is \textit{extractive},\footnote{An \textit{extractive} algorithm is one that can only generate a span from the given input passage $T$.}
and also allows for \textit{exact}\footnote{A decoding algorithm is considered \textit{exact} if it always generates the most likely sequence, as defined by the underlying model $P$, i.e. $\argmax_{a} P(a|T,Q)$.}
decoding since it computes a score for every possible span.

However, a recent trend in NLP is to frame all tasks as text-to-text \cite{2020t5, brown2020language}.
Indeed, various conditional language models have shown strong results on extractive QA \cite{2020t5,lewis-etal-2020-bart,chada2021fewshotqa}, generating answers via greedy decoding and its variants.
But how often does greedy decoding violate these properties \textit{in practice}, and does it actually affect its performance?

\section{The Exact-Extract Algorithm}

To study the greedy decoding algorithm in the context of extractive QA, we compare it to a new algorithm that produces the optimal extractive decoding (i.e. the span with the highest probability) from an autoregressive model: \textit{exact-extract}.

A naive optimal decoder can calculate the probability of every span $a=T_{i:i+j}$ individually.\footnote{We use Python-style span notations, i.e. zero-based indexing and exclusive boundaries. For example, $T_{2:4}$ refers to the span containing the third ($T_2$) and fourth ($T_3$) tokens.}
Using parallel computation hardware, this would require processing a batch of $n^2$ spans of up to length $n$ (where $n = |T|$), resulting in $O(n^3)$ space complexity.
In contrast, exact-extract uses dynamic programming to efficiently perform the same computation, with only $O(n^2)$ complexity.

The exact-extract algorithm is based on the observation that every span $a = T_{i:i+j}$ is the $j$th prefix of the suffix $T_{i:n}$.
Thus, for each suffix $T_{i:n}$, we can compute the probability of all of its prefixes $T_{i:i+j}$ in a single decoder forward pass. This process allows to calculate the probability of all possible spans, and select the one with the highest probability, making the algorithm both \textit{extractive} and \textit{exact}.

We now turn to a formal description of the algorithm. Let $\ell(\cdot, \cdot)$ and $e(\cdot, \cdot)$ denote local log-probabilities induced by the model $P$:\footnote{For clarity of notation, we assume that $P$ is always conditioned on $T$ and $Q$ as well, i.e. $P(x) = P(x|T,Q)$.} \begin{align*}
\ell (i, k) &= \log P(T_{i+k} | T_{i:i+k}) \\
e (i, k) &= \log P(\texttt{eos} | T_{i:i+k})
\end{align*}
Here, $\ell(i, k)$ is the log-probability of predicting the $k$-th token in the suffix $T_{i:n}$ (given its prefix $T_{i:i+k}$, à la teacher forcing), while $e(i, k)$ is the log-probability of ending the generated sequence at this point. 

For a fixed $i$, both $\ell(i,\cdot)$ and $e(i,\cdot)$ are calculated in a single decoder forward pass, as we simply "pool" the log-probability of the next token and the \texttt{eos} token for each prediction. We therefore need only $n$ decoder passes in order to derive $\ell(i,j)$ and $e(i,j)$ for all $i,j$.

Next, we denote the \textit{cumulative} log-probability of a sequence using $L(\cdot, \cdot)$:
\begin{align*}
L (i, j) &=  \log P(T_{i:i+j}) \\ 
&= \ell(i, 0) + \dots + \ell(i, j-1)
\end{align*}
In exact-extract, this value is dynamically calculated using a recursive formula:
\begin{align*}
L(i,0) &= 0 \\
L(i,j) &= L(i,j-1) + \ell(i,j-1)
\end{align*} 
At this point, $L(i,j)$ does not take into account the probability of generating the \texttt{eos} token. 
To derive the span's probability of being the answer, we sum the corresponding cumulative log-probability $L(\cdot, \cdot)$ with that of ending the sequence with an \texttt{eos} token $e(\cdot, \cdot)$:
\begin{align*}
\log &P(a = T_{i:i+j}) \nonumber \\ 
&=\log P(T_{i:i+j}) + \log P(\texttt{eos}|T_{i:i+j})  \nonumber\\
&= L(i,j) + e(i,j)
\end{align*}
Once we calculate $L(i,j)$ and $e(i,j)$ for all $(i,j)$, we retrieve the most probable span $T_{i:i+j}$ via:
\begin{align*}
    i , j = \argmax_{i,j} \big(L(i,j) + e(i, j)\big)
\end{align*}
Note that $L(\cdot, \cdot)$ is calculated directly from $\ell(\cdot, \cdot)$, and together with $e(\cdot, \cdot)$, we can derive the log-probability for all possible spans. Thus, $n$ decoder passes are sufficient for exact-extract, instead of $n^2$ passes required by the naive optimal decoder.

\begin{table*}[t!]
\small
\centering
\begin{tabular}{@{}llccccccccc@{}}
\toprule
\multirow{2}{*}{\textbf{Dataset}} &
  \textbf{Decoding} & \multicolumn{9}{c}{\textbf{\#Examples}} \\
& \textbf{Algorithm} &
  {\textbf{0}} &
  {\textbf{16}} &
  {\textbf{32}} &
  {\textbf{64}} &
  {\textbf{128}} &
  {\textbf{256}} &
  {\textbf{512}} &
  {\textbf{1024}} &
  \textbf{All} \\ 
\toprule

\multirow{2}{*}{\textbf{SQuAD}} & \textit{Greedy} &   50.4 &   81.3 &   84.1 &   86.0 &   88.3 &   89.0 &   90.3 &   \textbf{91.2} &   \textbf{94.5} \\
           & \textit{Exact-Extract} &   \textbf{60.0} &   \textbf{82.6} &   \textbf{85.2} &   \textbf{86.7} &   \textbf{89.0} &   \textbf{89.5} &   \textbf{90.5} &   \textbf{91.2} &   94.4 \\
\midrule
\multirow{2}{*}{\textbf{TriviaQA}} & \textit{Greedy} &   61.7 &   70.6 &   67.8 &   67.7 &   70.5 &   73.4 &   76.7 &   79.9 &   82.8 \\
           & \textit{Exact-Extract} &   \textbf{67.9} &   \textbf{74.8} &   \textbf{74.8} &   \textbf{75.3} &   \textbf{76.7} &   \textbf{77.6} &   \textbf{79.0} &   \textbf{80.5} &   \textbf{83.4} \\
\midrule
\multirow{2}{*}{\textbf{NaturalQs}} & \textit{Greedy} &   42.1 &   61.4 &   63.8 &   65.5 &   67.8 &   69.6 &   71.2 &   72.4 &   81.0 \\
           & \textit{Exact-Extract} &   \textbf{55.4} &   \textbf{64.4} &   \textbf{66.7} &   \textbf{68.5} &   \textbf{69.9} &   \textbf{71.2} &   \textbf{72.9} &   \textbf{73.6} &   \textbf{81.7} \\
\midrule
\multirow{2}{*}{\textbf{NewsQA}} & \textit{Greedy} &   19.2 &   41.7 &   45.3 &   45.3 &   48.0 &   51.6 &   56.3 &   61.4 &   71.0 \\
           & \textit{Exact-Extract} &   \textbf{36.3} &   \textbf{44.7} &   \textbf{48.8} &   \textbf{49.9} &   \textbf{51.8} &   \textbf{55.2} &   \textbf{58.3} &   \textbf{62.3} &   \textbf{71.8} \\
\midrule
\multirow{2}{*}{\textbf{SearchQA}} & \textit{Greedy} &   24.0 &   61.9 &   61.8 &   69.4 &   71.3 &   77.7 &   80.4 &   \textbf{83.0} &   \textbf{87.8} \\
           & \textit{Exact-Extract} &   \textbf{34.7} &   \textbf{64.1} &   \textbf{66.2} &   \textbf{71.7} &   \textbf{73.4} &   \textbf{78.9} &   \textbf{80.8} &   82.9 &   87.6 \\
\midrule
\multirow{2}{*}{\textbf{HotpotQA}} & \textit{Greedy} &   43.3 &   \textbf{66.3} &   \textbf{70.3} &   \textbf{73.1} &   \textbf{74.6} &   \textbf{76.4} &   \textbf{77.4} &   \textbf{78.7} &   \textbf{83.0} \\
           & \textit{Exact-Extract} &   \textbf{51.3} &   65.9 &   69.7 &   72.7 &   74.3 &   75.9 &   76.8 &   78.3 &   82.1 \\
\midrule
\multirow{2}{*}{\textbf{BioASQ}} & \textit{Greedy} &   55.5 &   \textbf{74.7} &   \textbf{76.8} &   \textbf{80.4} &   \textbf{85.2} &   \textbf{89.9} &   \textbf{92.2} &   \textbf{94.2} &     -- \\
           & \textit{Exact-Extract} &   \textbf{62.8} &   73.8 &   76.4 &   80.1 &   83.9 &   88.9 &   91.3 &   93.3 &     -- \\
\midrule
\multirow{2}{*}{\textbf{TextbookQA}} & \textit{Greedy} &   17.8 &   41.6 &   42.6 &   47.5 &   52.3 &   60.0 &   70.0 &   \textbf{73.5} &     -- \\
           & \textit{Exact-Extract} &   \textbf{36.0} &   \textbf{49.9} &   \textbf{51.2} &   \textbf{55.6} &   \textbf{58.0} &   \textbf{62.6} &   \textbf{70.8} &   73.4 &     -- \\
\bottomrule
\end{tabular}

\caption{Performance (F1) across all datasets and training set sizes of the few-shot QA benchmark, as well as the zero-shot setting (\textbf{0} examples, no fine-tuning), and the full-data setting (\textbf{all} examples) as in the 2019 MRQA Shared Task, containing an order of 100,000 training examples per dataset.} 
\label{tab:small_results_table}
\end{table*}

\section{Experimental Setup}
\label{sec:experimental_setup}
To measure how far from optimal is greedy decoding in practice,
we compare the performance of exact-extract and greedy decoding on a comprehensive few-shot QA benchmark.
\paragraph{Model}
We use T5-v1.1 \cite{2020t5, roberts-etal-2020-much}, an encoder-decoder transformer model pretrained to generate multiple randomly-masked spans.

We choose the v1.1 model checkpoint to avoid data contamination, as it was trained without any labeled data, while the original T5 models were trained in a multitask setting.
Our main experiments use T5-large (800M parameters).
To corroborate our findings are consistent across model sizes, we also measure the performance of T5-base (250M parameters).

\paragraph{Prompt}
Following the recent introduction of prompts for few-shot learning \cite{schick-schutze-2021-exploiting, schick-schutze-2021-just, gao-etal-2021-making, le-scao-rush-2021-many}, we align the task of extractive QA with T5's pretraining objective using a prompt. Specifically, the input to the encoder is:

\vspace{3pt}
\qquad Text: $T$

\qquad Question: $Q$

\qquad Answer:$\texttt{<extra\_id\_0>}$.
\vspace{3pt}

\noindent The model is trained to output:

\vspace{3pt}
\qquad $\texttt{<extra\_id\_0>}a\texttt{<extra\_id\_1>}$
\vspace{5pt}

Here, $T$ and $Q$ are the given passage and question, and $a$ is the expected answer.
This specific prompt was selected from 6 candidate prompts as part of the hyperparameter tuning process (see below), as recommended by \citet{Perez2021TrueFL}.\footnote{See Appendix~\ref{app:hp_search} for the full set of prompts.} 
\paragraph{Datasets}
We report results on the few-shot QA benchmark \cite{ram-etal-2021-shot}, created by subsampling 8 datasets from the MRQA 2019 shared task \cite{fisch-etal-2019-mrqa}: SQuAD \cite{rajpurkar-etal-2016-squad}, NewsQA \cite{trischler-etal-2017-newsqa}, TriviaQA \cite{joshi-etal-2017-triviaqa}, SearchQA \cite{dunn2017searchqa}, HotpotQA \cite{yang-etal-2018-hotpotqa}, Natural Questions \cite{kwiatkowski-etal-2019-natural}, BioASQ \cite{tsatsaronis_overview_2015}, and TextbookQA \cite{Kembhavi_2017_CVPR}.
Each dataset has a single fixed test set, and seven different training set sizes on a logarithmic scale from 16 to 1024 examples.
To account for sampling variation, five different training sets are sampled for each training set size, accumulating in 35 training sets for each of the 8 datasets.
For each dataset, we also examine the zero-shot setting (0 training examples) and the full-data setting (training on all examples). For BioASQ and TextbookQA, the largest setting we examine is 1024 examples, similar to \citet{ram-etal-2021-shot}.
Performance is measured via token-level F1 \cite{rajpurkar-etal-2016-squad} and averaged across the samples of each training set size.

\paragraph{Hyperparameters}
Hyperparameter tuning can be challenging in a few-shot setting because the development set (which needs to be taken out of an already-small training set) might have insufficient statistical power.
To address this issue, we assume that one ``academic'' dataset is available, which can provide enough validation examples for a modest hyperparameter search. 
The best hyperparameter configuration found via this single validation set is then used across all datasets and training sizes in our experiments. 
The academic dataset assumption follows the common practice of reusing hyperparameters tuned on larger data in prior work.

Specifically, we designate SQuAD as our academic dataset for hyperparameter tuning, and sample 2048 examples from its original training set to create a validation set.
We ensure that no example in the validation set contains a passage that appears in any of our few-shot training sets.
We then apply grid search on the following hyperparameters, for all 35 of SQuAD's few-shot training sets:
learning rate (1e-3, 2e-4, 1e-4, 5e-5), training steps (32, 64, 128, 256, 512, 1024, 2048), and prompts (see all 6 candidates in Appendix~\ref{app:hp_search}).
We select the single hyperparameter setting that optimizes performance across all training set sizes, as described in Appendix~\ref{app:hp_selection}.
This process yielded a learning rate of 5e-5 and 512 training steps, as well as the prompt described above.
Besides the tuned hyperparameters, we use the Adafactor optimizer \cite{Shazeer2018AdafactorAL}, a fixed batch size of 32,\footnote{For 16 training examples we use a batch size of 16.} and a dropout rate of 0.1.
This hyperparameter setting was applied universally to every dataset and data size in our experiments.

We did not perform any additional hyperparameter search for the full training set setups. Instead, we use the same hyperparameters selected for the few shot setting. A single exception is the number of epochs, which is set to 3 for all datasets.

\begin{figure*}
\centering
        \begin{subfigure}[t]{0.48\columnwidth}
                \centering
                \includegraphics[width=\linewidth]{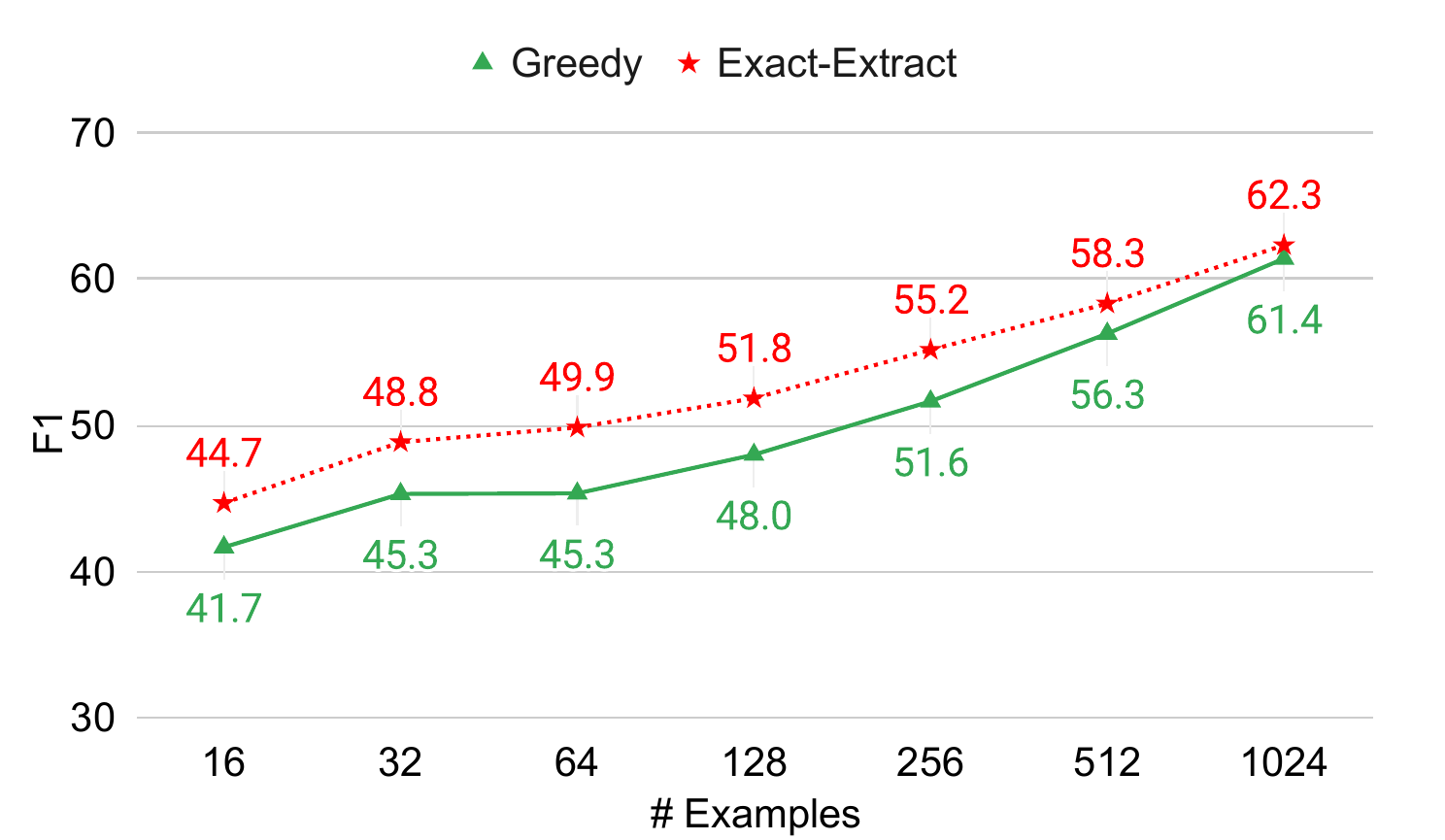}
                \caption{NewsQA}
                \label{fig:newsqa}
        \end{subfigure}
        \hfill
        \begin{subfigure}[t]{0.48\columnwidth}
                \centering
                \includegraphics[width=\linewidth]{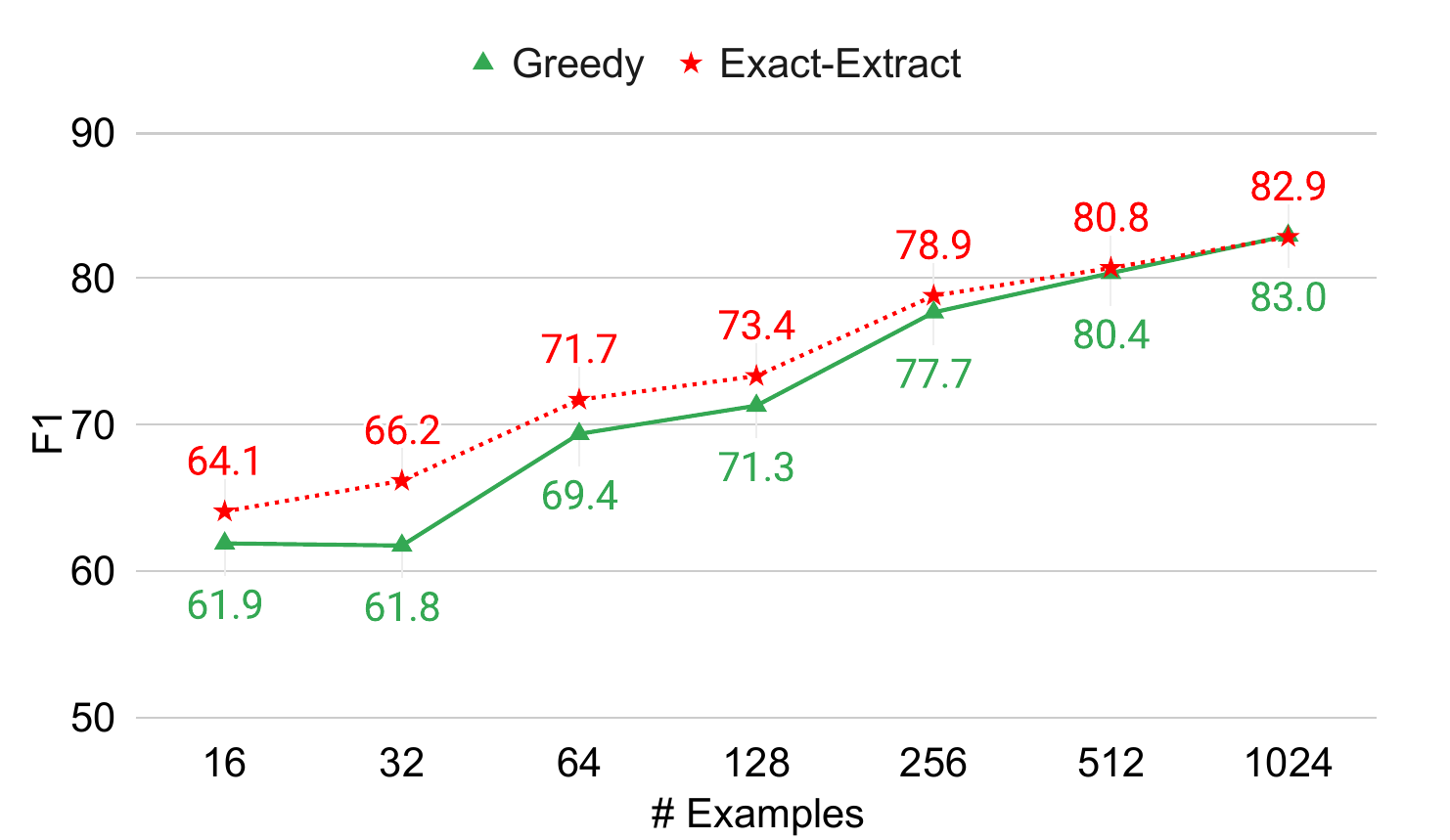}
                \caption{SearchQA}
                \label{fig:searchqa}
        \end{subfigure}
        \caption{Few-shot Performance (F1) of greedy decoding and exact-extract on NewsQA and SearchQA.}
        \label{fig:nq_newsqa}
\end{figure*}

\section{Results}

We first compare the performance of greedy decoding and exact-extract on the few-shot QA benchmark \cite{ram-etal-2021-shot}.
We observe the gap in performance consistently narrows as the training set get larger. When using 1024 training examples per dataset, greedy decoding lags only 0.3 points behind exact-extract on average.
We then show that greedy decoding becomes more extractive (and even more exact) as the training set increases in size, in line with the narrowing gap in performance.

\subsection{Performance}
\label{sec:greedy_vs_exact_extract}

Table~\ref{tab:small_results_table} shows our main performance results, covering all scenarios from zero-shot learning (0 examples) through few-shot learning (16 to 1024 examples) to the full-data setting (an order of 100,000 examples per dataset). The largest difference in performance is observed in the zero-shot setting, when no training examples are used. There, the advantage of exact-extract over greedy is substantial, with margins ranging from 6.2 points (TriviaQA) up to 18.2 (TextbookQA).
The large gaps across all datasets in the zero-shot setting suggest that when no task-specific training data is available, enforcing extractiveness and exactness through the decoding algorithm can greatly improve performance.

Nevertheless, when some annotated data \textit{is} available, the gap between greedy decoding and exact-extract shrinks at a dramatic pace.
Figure~\ref{fig:nq_newsqa} visualizes how increasing the training set closes the gap between the two decoding algorithms on NewsQA and SearchQA.
We observe that even 16 examples are sufficient to shrink the large gaps in the zero-shot setting to more modest, single-digit gaps, such as 3.0 points on NewsQA and 2.2 points on SearchQA (compared to 17.1- and 10.7-point gaps in the zero-shot setting, respectively).
Besides narrowing the performance gap, the shift from 0 to 16 labeled examples also results in a large \textit{absolute} improvement in performance, for both algorithms; in SQuAD, for instance, 16 examples are enough for the model to surpass the 80-point threshold. 

As the number of examples increase and reach 1024 and beyond (the full dataset), we observe that the performance difference between the two decoding algorithms diminishes, with less than one point separating the two, not necessarily in exact-extract's favor.

These trends are rather consistent across all datasets.
One notable anomaly is the small but consistent advantage of greedy decoding in the BioASQ and HotpotQA datasets. 
These datasets suffer from tokenization artifacts, which are particularly adversarial for exact-extract.
We analyze this phenomenon in depth in Appendix~\ref{sec:error_analysis}, and explain how the greedy algorithm's \textit{lack} of formal constraints can actually make it more robust to such issues.

We repeat our experiment using T5-base to verify that the observed trends are robust with respect to model size.
The full results of this experiment are available in Appendix~\ref{appx:base}.
Indeed, the main trend -- in which the performance difference between exact-extract and greedy decoding diminishes as more training examples become available -- emerges for the base model as well.

Finally, we compare greedy decoding with T5 to another extractive (and exact) system: Splinter \cite{ram-etal-2021-shot}. Splinter is an encoder-only transformer pretrained on heuristically-generated pseudo-questions, and has shown strong results on the few-shot QA benchmark.
+The comparison to Splinter is problematic due to different model sizes and pretraining corpora, but T5's overwhelmingly stronger results do provide yet another signal that the generative approach can be competitive, even when the decoding algorithm has no theoretical guarantees.
Detailed results are available in Appendix~\ref{appx:splinter}.

\subsection{How Extractive and Exact is Greedy?}
\label{sec:greedy_exact_extract_properties}

In Section~\ref{sec:greedy_vs_exact_extract} we observe that exact-exact substantially outperforms greedy decoding when no training examples are available, but that this gap quickly closes as more examples are added.
We hypothesize that the model acquires certain biases during fine-tuning, causing greedy decoding to produce more extractive and exact outputs.
We test our hypothesis by directly measuring both the \textit{extractiveness} and the \textit{exactness} of greedy decoding across different training set sizes.
Table~\ref{tab:copy_bias_match_exact_extract} shows the results.

\begin{table*}[t!]
\small
\centering
\begin{tabular}{@{}llccccccccc@{}}
\toprule
\multirow{2}{*}{\textbf{Dataset}} &
  {\multirow{2}{*}{\textbf{Metric}}} & \multicolumn{9}{c}{\textbf{\#Examples}} \\
&  &
  {\textbf{0}} &
  {\textbf{16}} &
  {\textbf{32}} &
  {\textbf{64}} &
  {\textbf{128}} &
  {\textbf{256}} &
  {\textbf{512}} &
  {\textbf{1024}} &
  \textbf{All} \\ 
 \midrule
\multirow{2}{*}{\textbf{SQuAD}} & \textit{Extract} &   33.1 &   87.4 &   86.0 &   89.2 &   92.1 &   92.7 &   93.9 &   95.3 &   99.5 \\
           & \textit{Exact} &   28.7 &   82.0 &   81.5 &   84.4 &   87.2 &   87.6 &   88.7 &   89.8 &   92.2 \\
\midrule
\multirow{2}{*}{\textbf{TriviaQA}} & \textit{Extract} &   68.7 &   87.6 &   84.8 &   83.7 &   85.5 &   88.6 &   91.3 &   94.2 &   92.7 \\
           & \textit{Exact} &   65.6 &   84.7 &   82.1 &   80.8 &   82.7 &   85.7 &   88.4 &   91.3 &   89.2 \\
\midrule
\multirow{2}{*}{\textbf{NaturalQs}} & \textit{Extract} &   51.5 &   80.3 &   82.4 &   82.5 &   87.2 &   89.2 &   91.6 &   93.8 &   98.5 \\
           & \textit{Exact} &   42.3 &   78.3 &   80.8 &   80.4 &   85.0 &   86.5 &   88.4 &   90.4 &   94.0 \\
\midrule
\multirow{2}{*}{\textbf{NewsQA}} & \textit{Extract} &   22.8 &   60.0 &   62.4 &   58.5 &   61.5 &   68.1 &   76.0 &   86.0 &   96.6 \\
           & \textit{Exact} &   21.2 &   55.8 &   58.9 &   54.9 &   57.9 &   64.5 &   70.8 &   79.2 &   91.1 \\
\midrule
\multirow{2}{*}{\textbf{SearchQA}} & \textit{Extract} &   44.9 &   83.8 &   79.0 &   83.6 &   84.4 &   86.9 &   90.0 &   92.5 &   90.9 \\
           & \textit{Exact} &   43.6 &   81.6 &   77.0 &   81.4 &   82.4 &   85.0 &   88.1 &   90.5 &   88.1 \\
\midrule
\multirow{2}{*}{\textbf{HotpotQA}} & \textit{Extract} &   60.8 &   89.9 &   91.6 &   94.0 &   95.4 &   96.0 &   96.8 &   97.3 &   99.6 \\
           & \textit{Exact} &   52.8 &   84.5 &   86.0 &   88.4 &   89.9 &   90.1 &   90.8 &   91.1 &   92.5 \\
\midrule
\multirow{2}{*}{\textbf{BioASQ}} & \textit{Extract} &   48.2 &   89.1 &   89.1 &   88.6 &   89.3 &   90.5 &   92.8 &   93.8 &     -- \\
           & \textit{Exact} &   43.2 &   85.9 &   85.7 &   85.8 &   86.0 &   87.5 &   89.7 &   91.0 &     -- \\
\midrule
\multirow{2}{*}{\textbf{TextbookQA}} & \textit{Extract} &   26.2 &   70.5 &   67.8 &   71.1 &   72.1 &   76.8 &   79.5 &   82.2 &     -- \\
           & \textit{Exact} &   21.2 &   65.5 &   63.8 &   67.9 &   68.5 &   72.6 &   75.4 &   77.8 &     -- \\
\bottomrule
\end{tabular}

\caption{Extractiveness and exactness of greedy decoding for all training set sizes. Extractiveness is the percentage of generated answers appearing in the passage. Exactness is the percentage of  generated texts identical to exact-extract output.}

\label{tab:copy_bias_match_exact_extract}
\end{table*}

\paragraph{Extractiveness}
We measure extractiveness as the percentage of examples for which greedy decoding generated a contiguous substring from the given passage.\footnote{For the sake of this analysis, we only count generated sequences that contain at least one alphanumeric character. Punctuation-only outputs (e.g. \textit{``.''}) are counted as not-extracted even though they appear in the context. These are common in the zero-shot setting.}
Table~\ref{tab:copy_bias_match_exact_extract} shows a steep increase in extractiveness when comparing 0 examples to 16.
In SQuAD for example, generating without any fine-tuning (zero-shot) results in only 33.1\% extractive outputs, whereas 16 training examples are enough to increase extractiveness to 87.4\%.
Extractiveness continues to increase as more examples are available, reaching nearly 100\% when training on the full dataset.
Effectively, the model acquires a \textit{copy bias} from training on labeled examples, which highly correlates with the increase in performance observed in (Table~\ref{tab:small_results_table}).

\paragraph{Exactness}
We measure exactness as the percentage of examples for which  greedy decoding produces the same output produced by exact-extract.

Table~\ref{tab:copy_bias_match_exact_extract} shows that there is a significant increase in the two algorithms' agreement rate as we introduce training examples.
However, unlike extractiveness, exactness does not reach nearly 100\%.
One possible explanation is that greedy decoding sometimes generates longer, yet just as correct, sequences in practice (i.e. greedy outputs "the IRA" while exact-extract outputs "IRA"). This is supported by our finding in Appendix~\ref{sec:greedy_errors}, where we show the model sometimes produces correct answers that differ from the ones annotated.

\begin{table*}[t]
\footnotesize
\centering
\begin{tabular}{@{}lllllllll@{}}
\toprule
  \textbf{Model} &
  \textbf{SQuAD} &
  \textbf{TriviaQA} & 
  \textbf{NQ} &
  \textbf{NewsQA} &
  \textbf{SearchQA} &
  \textbf{HotpotQA} &
  \textbf{BioASQ} &
  \textbf{TBQA} 
  \\ 
\midrule
\textit{Greedy}      & 
                    50.4 \textit{(33)} &  61.7 \textit{(69)} &        42.1 \textit{(51)} &  19.2 \textit{(23)} &  24.0 \textit{(45)} &  43.3 \textit{(61)} &  55.5 \textit{(48)} &  17.8 \textit{(26)} 
                    \\
\textit{~~+ RSS} & 
    \textbf{71.4} \textit{(61)} &  \textbf{69.3} \textit{(92)} &        57.2 \textit{(85)} &  \textbf{43.2} \textit{(78)} &  29.7 \textit{(74)} &  \textbf{59.0} \textit{(90)} &  65.5 \textit{(80)} &  39.0 \textit{(72)}
    \\ 
\midrule
\textit{Exact-Extract}
     & 60.0 &       67.9 &             55.4 &       36.3 &       34.7 &       51.3 &       62.8 &       36.0
    \\
\textit{~~+ RSS}
    & 69.4 
    & 67.8 
    & \textbf{58.1} 
    & 41.0 
    & \textbf{35.6} 
    & 57.1 
    & \textbf{66.9} 
    & \textbf{42.7} 
    \\
\bottomrule
\end{tabular}
\caption{\textbf{Top:} Zero-shot performance and extractiveness (in parentheses) of greedy decoding, with and without the RSS pretraining phase.
When no labeled examples are available, RSS pretraining greatly boosts both performance and extractiveness.
\textbf{Bottom:} Zero-shot performance of exact-extract, with and without the RSS pretraining phase.}
\label{tab:rss-zero-shot}
\end{table*}

\begin{figure*}[t]
\centering
\includegraphics[width=0.48\columnwidth]{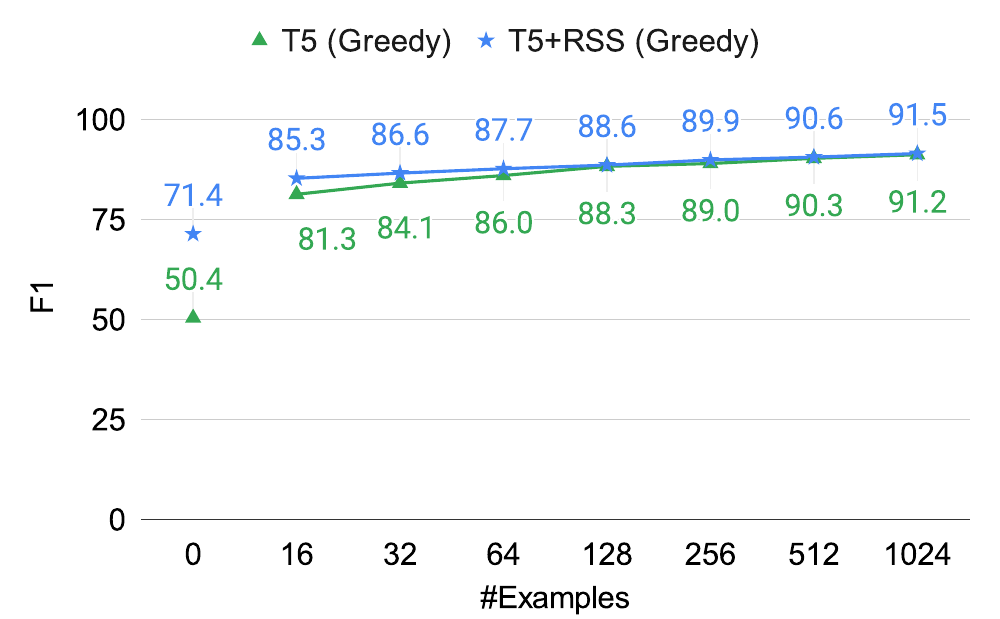}
\hfill
\includegraphics[width=0.48\columnwidth]{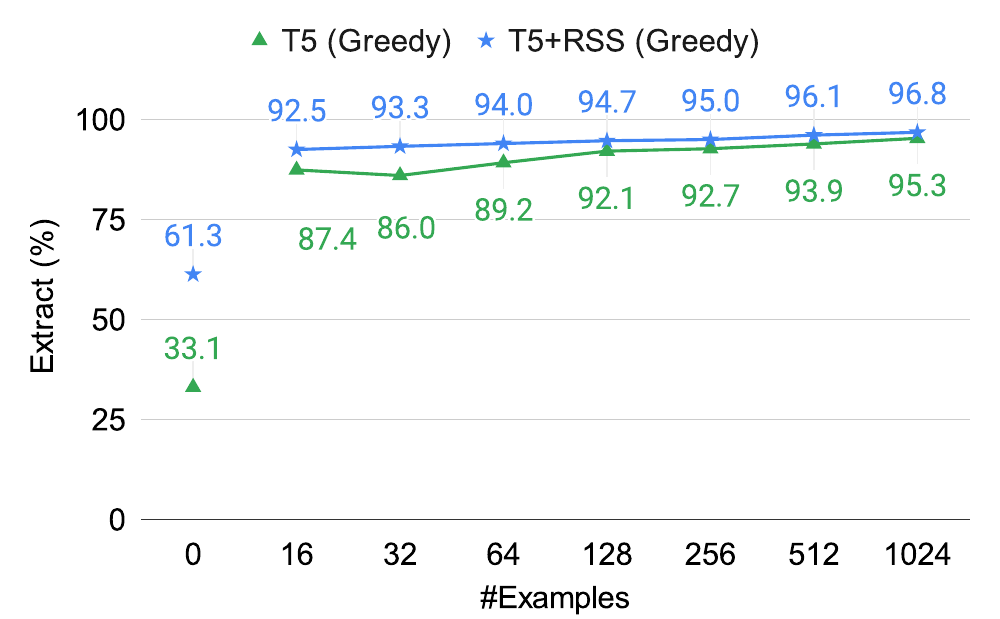}
\caption{\textbf{Left:} Performance of greedy decoding on SQuAD in zero-shot and few-shot settings, with and without the RSS pretraining phase. \textbf{Right:} Extractiveness of greedy decoding under the same settings. Performance of exact-extract in these settings is presented in Table~\ref{tab:rss-squad}.}
\label{fig:rss_squad}
\end{figure*}

\section{Pretraining Models to Extract}

In Section~\ref{sec:greedy_exact_extract_properties} we observe a strong correlation between performance and a model's tendency to generate answers extracted from the context.
Can we make a model more extractive via pretraining? 

Inspired by recent work on pretraining encoders for span selection, we propose applying an additional pretraining phase (mid-training) to T5 before fine-tuning.
We adapt the recurring span selection objective (RSS) used in Splinter \cite{ram-etal-2021-shot,ram-etal-2022-learning} to the generative setting: (1) find non-stopword spans that occur more than once in a given passage, (2) mask one instance of a recurring span, (3) train the model to predict the original content of the masked span.
While Splinter is trained by masking multiple different spans in parallel, we limit ourselves to a single span in each passage to better approximate the target task.
For this experiment, we create 100,000 RSS pretraining examples from English Wikipedia, using WikiExtractor \cite{Wikiextractor2015}. We pretrain T5-large on this dataset for 3 epochs.\footnote{The trained model is available via the Transformers library \cite{wolf-etal-2020-transformers}:  \url{https://huggingface.co/tau/t5-v1_1-large-rss}}
For simplicity, we use the same hyperparameter configuration from Section~\ref{sec:experimental_setup}.

Table~\ref{tab:rss-zero-shot} shows that incorporating RSS pretraining substantially boosts the extractiveness of greedy decoding in the zero-shot setting, as well as its performance.
Exact-extract also benefits from RSS pretraining (but the relative performance gains are smaller), even though it is already 100\% extractive. Therefore, we hypothesize that RSS pretraining encourages additional properties that benefit extractive question answering, beyond just copying.

That being said, the advantage of adding an RSS pretraining phase wanes as more labeled examples are available, even when greedy decoding is used.
Figure~\ref{fig:rss_squad} shows how the original T5 model quickly catches up on the RSS-pretrained model's performance on SQuAD.
Notably, when using 128 or more labeled examples, the benefit from adding RSS pretraining is less than one F1 point.
This behavior is somewhat expected given our observations in Section~\ref{sec:greedy_exact_extract_properties}, where we observe a steep rise in both extractiveness and performance once annotated examples are introduced. Hence, adding labeled examples might be more consequential then adding an RSS pretraining phase.

\section{Error Analysis}
\label{sec:error_analysis}
\begin{table*}[t]
\small
\centering
\begin{tabular}{@{}lllllll@{}}
\toprule
  \textbf{Test Subset} &
  \textbf{SQuAD} &
  \textbf{TriviaQA} & 
  \textbf{SearchQA} &
  \textbf{HotpotQA} &
  \textbf{BioASQ} &
  \textbf{TextbookQA} 
  \\ 
\midrule
$S_{out}$ & 
                    75.2 \textit{~~(3\%)} & 
                    39.9 \textit{~~(2\%)} & 
                    72.2 \textit{~~(9\%)} & 
                    64.2 \textit{~~(6\%)} & 
                    59.0 \textit{~~(6\%)} & 
                    46.7 \textit{~~(2\%)}   
                    \\
$S_{in}$ & 
                    91.6 \textit{(97\%)} & 
                    81.2 \textit{(98\%)} & 
                    84.0 \textit{(91\%)} & 
                    79.1 \textit{(94\%)} & 
                    95.6 \textit{(94\%)} & 
                    
                    74.1 \textit{(98\%)}   
                    \\
\bottomrule
\end{tabular}
\caption{Performance of exact-extract on two complementary test set subsets: $S_{out}$ and $S_{in}$. An $S_{out}$ subset contains only examples in which the tokenized answer is \textit{not} a subsequence of the tokenized passage. An $S_{in}$ subset contains the rest of the test set examples. The relative size of each subset appears in parentheses. NaturalQuestions and NewsQA are omitted from this table since their test sets are 100\% extractive. Models are the same used to report results on 1024 training examples in Table~\ref{tab:small_results_table}.
}
\label{tab:tokenization}
\end{table*}

In theory, exact-extract is an optimal decoding algorithm.
However, the results in Section~\ref{sec:greedy_vs_exact_extract} show that greedy decoding sometimes performs better than exact-extract in practice.
Analyzing these cases reveals that inconsistent tokenization can cause the annotated answer to become non-extractive, deteriorating the performance of exact-extract (Section~\ref{sec:tokenization}).
We then analyze the greedy algorithm's errors, and observe that almost half the errors are correct answers, even if not always extractive.

\subsection{Exact-Extract}
\label{sec:tokenization}

In some datasets, such as BioASQ and HotpotQA, we observe that the greedy algorithm performs better on average than exact-extract (see Table~\ref{tab:small_results_table}).
A manual analysis reveals that often in these cases the \textit{tokenized} annotated answer is not a subsequence of the \textit{tokenized} passage.
For example, a passage containing the text \textit{``(1971)''} is tokenized as [\textit{``\_(19'', ``71'', ``)''}], while the answer string \textit{``1971''} is tokenized as [\textit{``\_1971''}].

To measure the prevalence and effect of this phenomenon, we partition each test set into two: $S_{out}$ and $S_{in}$. $S_{out}$ subsets include all test examples where the tokenized answer is \textit{not} a subsequence of the tokenized passage. $S_{in}$ subsets include the rest of the test set.
Then, for each model from our main experiment (Section~\ref{sec:greedy_vs_exact_extract}), we measure the performance of exact-extract on $S_{out}$ and $S_{in}$.

Table~\ref{tab:tokenization} shows that exact-extract performs substantially worse on $S_{out}$ subsets. This is expected, as they are designed to contain only answers which cannot be extracted (token-wise) from the passage. In addition, we observe that in the datasets where exact-exact was outperformed by greedy, $S_{out}$ is relatively larger compared to $S_{in}$.

The tokenization issue behind this phenomenon stems from the way subword token vocabularies are commonly induced \cite{sennrich-etal-2016-neural, kudo-richardson-2018-sentencepiece}.
It is quite likely that this phenomenon disappears when using character-level or byte-level tokenization \cite{shaham-levy-2021-neural, xue2021byt5}.
However, the fact that greedy decoding is \textit{not} 100\% extractive actually allows it to overcome tokenization mismatches and generate the annotated answer.

\subsection{Greedy Decoding}
\label{sec:greedy_errors}

We analyze the cases in which exact-extract did produce the annotated answer, but the greedy algorithm did not.
This allows us to decouple the model from the decoding algorithm, since we know that the most likely span according to the model is indeed correct.
Specifically, we analyzed results from models trained on 1024 examples, sampling up to 20 examples from each dataset.
\begin{table}[t]
\small
\centering
\begin{tabular}{@{}lr@{}}
\toprule
\textbf{Category} & \textbf{Frequency} \\
\midrule
Incorrect Answer & 51.9\% \\
Correct Answer & 48.1\% \\
~~~~Annotation Error & 17.6\% \\
~~~~Not Extractive & 30.5\% \\
~~~~~~~~Paraphrase & 23.6\% \\
~~~~~~~~Added Information & 6.9\% \\
\bottomrule
\end{tabular}
\caption{Error analysis of greedy decoding, based on models trained on 1024 examples. All cases reflect examples where exact-extract accurately produced the annotated answer, while the greedy algorithm did not.}
\label{tab:greedy_failures}
\end{table}

Table~\ref{tab:greedy_failures} breaks down the errors into a hierarchy of categories, alongside the prevalence of each error type.
We observe that approximately half of the errors (48.1\%) are semantically correct answers.
Of those, about a third account for annotation errors, typically where there can be multiple correct spans but only one appeared in the test set (and the greedy algorithm chose another).

The other two thirds are particularly interesting: they are semantically correct, but on the other hand, they are not extractive.
The majority of these cases are paraphrases, where the model elaborates a bit more (annotated: ``shamed'', generated: ``he shamed him''), or replaces a number-word with the actual number (annotated: ``sixty percent'', generated: ``60\%'').
Most curiously, in about a quarter of the correct answers which are not extractive, the model adds information that was not mentioned in the original passage, e.g. generating ``Queen Elizabeth II'' instead of the span ``the Queen''.
In contrast with \textit{hallucination}, commonly reported in summarization tasks \cite{lewis-etal-2020-bart, zhao-etal-2020-reducing}, the information added answers is  correct.

One can debate whether non-extractive answers are actually correct.
On one hand, the task is defined as \textit{extractive} QA.
Having said that, these answers do fulfill a potential user's information need, and may even benefit said user by containing additional context.

\section{Conclusions}

We investigate the optimality of greedy decoding for extractive question answering by comparing it to \textit{exact-extract}, an optimal decoding algorithm that guarantees both \textit{extractiveness} and \textit{exactness}.
While the greedy algorithm lags behind exact-extract in the zero-shot setting, training the model on as few as 16 labeled examples shrinks the performance gap substantially.
This gap continues to narrow as more examples are available, typically converging to less than 1 point (F1) when training on 1024 examples. 
Overall, our results showcase the impressive ability of pretrained language models to adapt to extractive question answering while relying only on a naive decoding algorithm.

\section*{Acknowledgements}
We thank Mor Geva and anonymous reviewers
for valuable feedback and discussions.
This work was supported by the Tel Aviv University Data Science Center, Len Blavatnik and the Blavatnik Family foundation, the Alon Scholarship, Intel Corporation, and the Yandex Initiative for Machine Learning.

\bibliography{anthology,custom}
\bibliographystyle{plainnat}

\appendix
\label{app}
\section{Hyperparameter Search Space}
\label{app:hp_search}
Our search space includes three hyperparameters: learning rate, number of training steps and the prompt. We choose from the following candidate sets:
\begin{itemize}
    \item Learning rates: \{1e-3, 2e-4, 1e-4, 5e-5\} 
    \item Number of training steps: \{32, 64, ..., 2048\}
    \item Prompts: See Table~\ref{tab:prompts} for the list of prompts considered.
\end{itemize}

\noindent Following the hyperparameters selection process (see Appendix~\ref{app:hp_selection}), we proceed with a learning rate of 5e-5 and train for 512 steps, with the second prompt from Table~\ref{tab:prompts}.

\section{Hyperparameter Selection}
\label{app:hp_selection}
We describe our approach for selecting the best hyperparameter configuration.
As described in Section~\ref{sec:experimental_setup}, we use SQuAD's 35 training sets; 7 different sizes with 5 sets each, alongside a 2048-example validation set. 

Formally, denote the set of training sizes by $N=\{16,32,...,1024\}$ and the number of different sets for each size by $K$ ($K=5$ in our case). We define $s_i^{n,k}$ as the model performance on the validation set when trained on the $k$-th training set of of size $n\in N$, using the hyperparameter configuration $h_i$.\footnote{$h_i$ defines a specific learning rate, number of training steps and a prompt (see Appendix~\ref{app:hp_search}).}  Following, we take $s_i^n$ to be the score of $h_i$ averaged across datasets of size $n$, i.e:
\begin{align*}
 s_i^n = \frac{1}{K}\sum_{k=1}^K s_i^{n,k}   
\end{align*}
Next, we normalize $s_i^n$ by the maximal averaged score on datasets of size $n$:
\begin{align*}
    \Tilde{s}_i^n = \frac{s_i^n}{\max_{j} s_j^n} 
\end{align*}
Finally, we average $h_i$'s normalized scores across sizes:
\begin{align*}
    s_i = \frac{1}{|N|}\sum_{n\in N } \Tilde{s}_i^n
\end{align*}
The hyperparameters configuration $h_{i^*}$ is chosen via $ i^* = \argmax_{i} s_i$.
\begin{table}[t]
    \centering
\begin{tabular}{@{}l@{}}
\toprule
$T$\\Question: $Q$\\Answer:<$\texttt{extra\_id\_0}$>. \\
\midrule
Text: $T$\\Question: $Q$\\Answer:<$\texttt{extra\_id\_0}$>. \\
\midrule
$T$\\$Q$\\<$\texttt{extra\_id\_0}$>. \\
\midrule
$T$\\Answer the following question based on the \\
above text: $Q$\\<$\texttt{extra\_id\_0}$>. \\
\midrule
Please read the following paragraph and answer \\
the question at the end:\\$T$\\$Q$\\<$\texttt{extra\_id\_0}$>. \\
\midrule
Background: $T$\\Q: $Q$\\\textit{A}:<$\texttt{extra\_id\_0}$> \\
\bottomrule
\vspace{-0.5cm}
\caption{Prompts considered during hyperparameter grid search. The placeholders $T$ and $Q$ are replaced with the example's passage and question, respectively; <$\texttt{extra\_id\_0}$> is T5's sentinel token representing a masked span.}
\label{tab:prompts}
\end{tabular}
\end{table}

\section{Results with T5-base}
\label{appx:base}

\begin{table*}[t!]
\small
\centering
\begin{tabular}{@{}llcccccccc@{}}
\toprule
\multirow{2}{*}{\textbf{Dataset}} &
  \textbf{Decoding} & \multicolumn{8}{c}{\textbf{\#Examples}} \\
& \textbf{Algorithm} &
  {\textbf{0}} &
  {\textbf{16}} &
  {\textbf{32}} &
  {\textbf{64}} &
  {\textbf{128}} &
  {\textbf{256}} &
  {\textbf{512}} &
  {\textbf{1024}} \\ 
\midrule
\multirow{2}{*}{\textbf{SQuAD}}      & \textit{Greedy}        & 29.7 & 50.7 & 53.3 & 60.3 & 69.6 & 72.8 & 76.4 & 76.5 \\ \
                                     & \textit{Exact-Extract} & \textbf{34.6} & \textbf{54.9} & \textbf{57.9} & \textbf{62.7} & \textbf{71.0} & \textbf{73.5} & \textbf{77.0} & \textbf{76.8} \\
\midrule
\multirow{2}{*}{\textbf{TriviaQA}}   & \textit{Greedy}        & 54.1 & 37.1 & 29.5 & 38.2 & 52.0 & 51.4 & 67.1 & \textbf{68.5} \\
                                     & \textit{Exact-Extract} & \textbf{54.5} & \textbf{50.9} & \textbf{48.0} & \textbf{51.7} & \textbf{58.3} & \textbf{54.8} & \textbf{67.5} & \textbf{68.5} \\
\midrule
\multirow{2}{*}{\textbf{NaturalQs}} & \textit{Greedy} & 13.9 & 35.1 & 39.7 & 44.1 & 50.0 & 52.1 & 54.3 & 55.2 \\
                                     & \textit{Exact-Extract} & \textbf{34.4} & \textbf{42.1} & \textbf{44.8} & \textbf{49.3} & \textbf{53.0} & \textbf{54.2} & \textbf{55.5} & \textbf{56.1} \\
\midrule
\multirow{2}{*}{\textbf{NewsQA}}     & \textit{Greedy}        & 25.0 & 20.7 & 22.3 & 26.2 & 34.3 & 39.6 & \textbf{42.4} & \textbf{44.1} \\
                                     & \textit{Exact-Extract} & \textbf{27.6} & \textbf{28.9} & \textbf{30.5} & \textbf{32.3} & \textbf{36.7} & \textbf{40.1} & 41.9 & 43.3 \\
\midrule
\multirow{2}{*}{\textbf{SearchQA}}   & \textit{Greedy}        & 4.8  & 29.5 & 27.6 & 38.1 & 51.7 & 59.7 & \textbf{65.2} & \textbf{64.3} \\
                                     & \textit{Exact-Extract} & \textbf{13.4} & \textbf{37.4} & \textbf{37.8} & \textbf{42.4} & \textbf{53.1} & \textbf{59.9} & \textbf{65.2} & 64.2 \\
\midrule
\multirow{2}{*}{\textbf{HotpotQA}}   & \textit{Greedy}        & 33.3 & 38.5 & 42.5 & 53.7 & 59.5 & \textbf{62.5} & \textbf{66.0} & \textbf{65.5} \\
                                     & \textit{Exact-Extract} & \textbf{41.2} & \textbf{40.9} & \textbf{44.8} & \textbf{54.5} & \textbf{59.6} & 62.3 & 65.5 & 64.6 \\
\midrule
\multirow{2}{*}{\textbf{BioASQ}}     & \textit{Greedy}        & 42.8 & 39.5 & 51.0 & 63.1 & \textbf{73.8} & 79.3 & 81.9 & \textbf{81.9} \\
                                     & \textit{Exact-Extract} & \textbf{46.5} & \textbf{42.3} & \textbf{52.0} & \textbf{64.2} & 72.9 & \textbf{79.4} & \textbf{82.1} & \textbf{81.9} \\
\midrule
\multirow{2}{*}{\textbf{TextbookQA}} & \textit{Greedy}        & 9.0  & 8.8  & 9.7  & 14.4 & 21.1 & 34.8 & 43.6 & \textbf{48.6} \\
                                     & \textit{Exact-Extract} & \textbf{18.9} & \textbf{17.2} & \textbf{18.7} & \textbf{19.9} & \textbf{24.9} & \textbf{36.2} & \textbf{43.9} & 48.1 \\
\bottomrule
\end{tabular}
\caption{Performance (F1) of T5-base across all datasets and training set sizes of the few-shot QA benchmark, as well as the zero-shot setting (\textbf{0} examples, no fine-tuning) as in the 2019 MRQA Shared Task.}
\label{tab:t5_base_results_table}
\end{table*}
\begin{table*}[t!]
\centering
\begin{tabular}{@{}lllllllll@{}}
\toprule
\multirow{2}{*}{\textbf{Model}} & \multicolumn{8}{c}{\textbf{\#Examples}} \\
  &
  {\textbf{0}} &
  {\textbf{16}} &
  {\textbf{32}} &
  {\textbf{64}} &
  {\textbf{128}} &
  {\textbf{256}} &
  {\textbf{512}} &
  {\textbf{1024}} \\ 
\midrule

\textit{Greedy}      & 
  50.4 &
  81.3 &
  84.1 &
  86.0 &
  88.3 &
  89.0 &
  90.3 &
  91.2 \\ 
  \textit{~~+RSS} &
  \textbf{71.4} &
  85.3 &
  86.6 &
  87.7 &
  88.6 &
  89.9 &
  90.6 &
  91.5 \\
  \midrule
  \textit{Exact-Extract}      &
  60.0 & 
  82.6 & 
  85.2 & 
  86.7 & 
  89.0 & 
  89.5 & 
  90.5 & 
  91.2
  \\
  \textit{~~+RSS} & 
  69.4 &	
  \textbf{85.6} &	
  \textbf{86.7} &	
  \textbf{87.9} &	
  \textbf{89.4} &	
  \textbf{90.2} &	
  \textbf{90.7} &	
  \textbf{91.9}
  \\
\bottomrule
\end{tabular}
\caption{Performance (F1) of T5-large on SQuAD across all training set sizes in few-shot QA benchmark, as well as the zero-shot setting (0 examples, no finetuning on the task). Performance is measured for both greedy decoding and exact-extract, with and without the RSS pretraining phase. Exact-Extract shows very little to no advantage over greedy decoding once sufficient amount of examples is available, both for greedy decoding and exact-extract.}
\label{tab:rss-squad}
\end{table*}
Table~\ref{tab:t5_base_results_table} shows performance results when using T5-base in the zero-shot setting and all few-shot settings. The trends are similar; the gap between exact-extract and greedy decoding narrows as more training examples are present.
\begin{table*}[t!]
\small
\centering
\begin{tabular}{@{}llcccccccc@{}}
\toprule
\multirow{2}{*}{\textbf{Dataset}} &
  {\multirow{2}{*}{\textbf{Model}}} & \multicolumn{8}{c}{\textbf{\#Examples}} \\
    &  &
  {\textbf{0}} &
  {\textbf{16}} &
  {\textbf{32}} &
  {\textbf{64}} &
  {\textbf{128}} &
  {\textbf{256}} &
  {\textbf{512}} &
  {\textbf{1024}} \\ 
\toprule
\multirow{3}{*}{\textbf{SQuAD}} &
  \textit{T5-large} &
  \textbf{50.4} &
  \textbf{81.3} &
  \textbf{84.1} &
  \textbf{86.0} &
  \textbf{88.3} &
  \textbf{89.0} &
  \textbf{90.3} &
  \textbf{91.2} \\
 &
  \textit{T5-base} &
  29.7 &
  50.7 &
  53.3 &
  60.3 &
  69.6 &
  72.8 &
  76.4 &
  76.5 \\
 &
  \textit{Splinter-large} &
  -- &
  -- &
  70.0 &
  75.8 &
  80.4 &
  81.9 &
  85.1 &
  86.3 \\ \midrule
\multirow{3}{*}{\textbf{TriviaQA}} &
  \textit{T5-large} &
  \textbf{61.7} &
  \textbf{70.6} &
  \textbf{67.8} &
  \textbf{67.7} &
  \textbf{70.5} &
  \textbf{73.4} &
  \textbf{76.7} &
  \textbf{79.9} \\
 &
  \textit{T5-base} &
  54.1 &
  37.1 &
  29.5 &
  38.2 &
  52.0 &
  51.4 &
  67.1 &
  68.5 \\
 &
  \textit{Splinter-large} &
  -- &
  -- &
  45.3 &
  55.3 &
  58.1 &
  66.1 &
  40.8 &
  71.0 \\ \midrule
\multirow{3}{*}{\textbf{NaturalQs}} &
  \textit{T5-large} &
  \textbf{42.1} &
  \textbf{61.4} &
  \textbf{63.8} &
  \textbf{65.5} &
  \textbf{67.8} &
  \textbf{69.6} &
  \textbf{71.2} &
  \textbf{72.4} \\
 &
  \textit{T5-base} &
  13.9 &
  35.1 &
  39.7 &
  44.1 &
  50.0 &
  52.1 &
  54.3 &
  55.2 \\
 &
  \textit{Splinter-large} &
  -- &
  -- &
  40.6 &
  46.3 &
  54.4 &
  48.8 &
  64.1 &
  67.9 \\ \midrule
\multirow{3}{*}{\textbf{NewsQA}} &
  \textit{T5-large} &
  19.2 &
  \textbf{41.7} &
  \textbf{45.3} &
  \textbf{45.3} &
  \textbf{48.0} &
  51.6 &
  56.3 &
  \textbf{61.4} \\
 &
  \textit{T5-base} &
  \textbf{25.0} &
  20.7 &
  22.3 &
  26.2 &
  34.3 &
  39.6 &
  42.4 &
  44.1 \\
 &
  \textit{Splinter-large} &
  -- &
  -- &
  33.7 &
  36.0 &
  47.7 &
  \textbf{52.3} &
  \textbf{57.4} &
  58.5 \\ \midrule
\multirow{3}{*}{\textbf{SearchQA}} &
  \textit{T5-large} &
  \textbf{24.0} &
  \textbf{61.9} &
  \textbf{61.8} &
  \textbf{69.4} &
  \textbf{71.3} &
  \textbf{77.7} &
  \textbf{80.4} &
  \textbf{83.0} \\
 &
  \textit{T5-base} &
  4.8 &
  29.5 &
  27.6 &
  38.1 &
  51.7 &
  59.7 &
  65.2 &
  64.3 \\
 &
  \textit{Splinter-large} &
  -- &
  -- &
  39.9 &
  42.0 &
  52.0 &
  60.7 &
  65.0 &
  68.5 \\ \midrule
\multirow{3}{*}{\textbf{HotpotQA}} &
  \textit{T5-large} &
  \textbf{43.3} &
  \textbf{66.3} &
  \textbf{70.3} &
  \textbf{73.1} &
  \textbf{74.6} &
  \textbf{76.4} &
  \textbf{77.4} &
  \textbf{78.7} \\
 &
  \textit{T5-base} &
  33.3 &
  38.5 &
  42.5 &
  53.7 &
  59.5 &
  62.5 &
  66.0 &
  65.5 \\
 &
  \textit{Splinter-large} &
  -- &
  -- &
  53.2 &
  60.5 &
  65.5 &
  55.7 &
  72.1 &
  74.1 \\ \midrule
\multirow{3}{*}{\textbf{BioASQ}} &
  \textit{T5-large} &
  \textbf{55.5} &
  \textbf{74.7} &
  \textbf{76.8} &
  \textbf{80.4} &
  \textbf{85.2} &
  \textbf{89.9} &
  \textbf{92.2} &
  \textbf{94.2} \\
 &
  \textit{T5-base} &
  42.8 &
  39.5 &
  51.0 &
  63.1 &
  73.8 &
  79.3 &
  81.9 &
  81.9 \\
 &
  \textit{Splinter-large} &
  -- &
  -- &
  58.8 &
  55.1 &
  77.0 &
  82.3 &
  86.7 &
  91.4 \\ \midrule
\multirow{3}{*}{\textbf{TextbookQA}} &
  \textit{T5-large} &
  \textbf{17.8} &
  \textbf{41.6} &
  \textbf{42.6} &
  47.5 &
  \textbf{52.3} &
  \textbf{60.0} &
  \textbf{70.0} &
  \textbf{73.5} \\
 &
  \textit{T5-base} &
  9.0 &
  8.8 &
  9.7 &
  14.4 &
  21.1 &
  34.8 &
  43.6 &
  48.6 \\
 &
  \textit{Splinter-large} &
  -- &
  -- &
  39.5 &
  \textbf{47.7} &
  52.2 &
  57.5 &
  49.7 &
  51.6 \\
\bottomrule
\end{tabular}
\caption{Performance (F1) of T5-large (greedy decoding), T5-base (greedy decoding) and Splinter-large \cite{ram-etal-2021-shot}, across all datasets and training set sizes of the few-shot QA benchmark, as well as the zero-shot setting (\textbf{0} examples, no fine-tuning), and the full-data setting (\textbf{all} examples) as in the 2019 MRQA Shared Task, containing an order of 100,000 training examples per dataset. Splinter-large results were available for 32 examples or more.}
\label{tab:t5_splinter_f1}
\end{table*}
\section{Comparison with Splinter}
\label{appx:splinter}
We present T5-large and T5-base greedy decoding results alongside those of Splinter-large\footnote{The results reported in \citet{ram-etal-2021-shot} were obtained using Splinter-base. The authors shared new results with us, obtained with Splinter-large.} in Table~\ref{tab:t5_splinter_f1}. 

Although the models cannot be fairly compared (due to different sizes, training corpora and duration of training), T5-large outperforms Splinter-large across all datasets and size regimes; the margin ranges from 14 F1 points on average for 16-64 examples, to 9 points for 128-1024 training examples.

\end{document}